\documentclass{article}

\usepackage[preprint]{neurips_2024}

\usepackage{times}          
\usepackage{helvet}         
\usepackage{courier}        
\usepackage[hyphens]{url}   
\usepackage{graphicx}       
\usepackage{caption}        
\usepackage{authblk}
\usepackage{wrapfig}
\usepackage[utf8]{inputenc} 
\usepackage[T1]{fontenc}    
\usepackage{microtype}      
\usepackage{hyperref}       

\usepackage{booktabs}       
\usepackage[xcdraw,table]{xcolor} 
\usepackage{multirow}       
\usepackage{epstopdf}       
\usepackage{float}          
\usepackage{subcaption}     
\usepackage{svg}            
\usepackage{url}
\usepackage{amsmath}        
\usepackage{amsthm}         
\usepackage{amssymb}        
\usepackage{amsfonts}       
\usepackage{bm}             
\usepackage{nicefrac}       
\usepackage[linesnumbered,ruled,vlined,boxed,noend]{algorithm2e} 

\usepackage{listings}       
\usepackage{newfloat}       
\newtheorem{definition}{Definition}

\title{A New Perspective on Privacy Protection in Federated Learning with Granular-Ball Computing}

\author[1]{Guannan Lai}
\author[1]{Yihui Feng}
\author[1]{Xin Yang\thanks{Corresponding author. Email: \texttt{yangxin@swufe.edu.cn}}\;\;}
\author[1]{Xiaoyu Deng}
\author[1]{Hao Yu}
\author[2]{Shuyin Xia}
\author[2]{Guoyin Wang}
\author[3]{Tianrui Li}

\affil[1]{Southwestern University of Finance and Economics, China \\\texttt{aignlai@163.com, fengyh0727@163.com, xydeng2809@163.com, yuhao2033@163.com}}
\affil[2]{Chongqing University of Posts and Telecommunications, China \\\texttt{xiasy@cqupt.edu.cn, wanggy@cqupt.edu.cn}}
\affil[3]{Southwest Jiaotong University, China \\\texttt{trli@swjtu.edu.cn}}

\begin{document}

\maketitle

\begin{abstract}
  Federated Learning (FL) facilitates collaborative model training while prioritizing privacy by avoiding direct data sharing. However, most existing articles attempt to address challenges within the model's internal parameters and corresponding outputs, while neglecting to solve them at the input level. To address this gap, we propose a novel framework called Granular-Ball Federated Learning (GrBFL) for image classification. GrBFL diverges from traditional methods that rely on the finest-grained input data. Instead, it segments images into multiple regions with optimal coarse granularity, which are then reconstructed into a graph structure. We designed a two-dimensional binary search segmentation algorithm based on variance constraints for GrBFL, which effectively removes redundant information while preserving key representative features. Extensive theoretical analysis and experiments demonstrate that GrBFL not only safeguards privacy and enhances efficiency but also maintains robust utility, consistently outperforming other state-of-the-art FL methods.
The code is available at \url{https://github.com/AIGNLAI/GrBFL}.

\end{abstract}

\section{Introduction}

Federated Learning (FL) is a distributed machine learning approach to solve the data silo problem, which aims to train a global model without the need to centralize the original dataset. The important problem of FL to be solved now is the trade-off between privacy, efficiency, and utility \cite{zhang2022no}, i.e., how to protect privacy without significantly compromising utility and efficiency. However, existing FL methods often focus on improving model performance from the perspective of models and features, with few methods considering whether changing the input data of FL can enhance the model's overall performance. 

\begin{figure}[ht]
  \centering
  \includegraphics[width=0.8\linewidth]{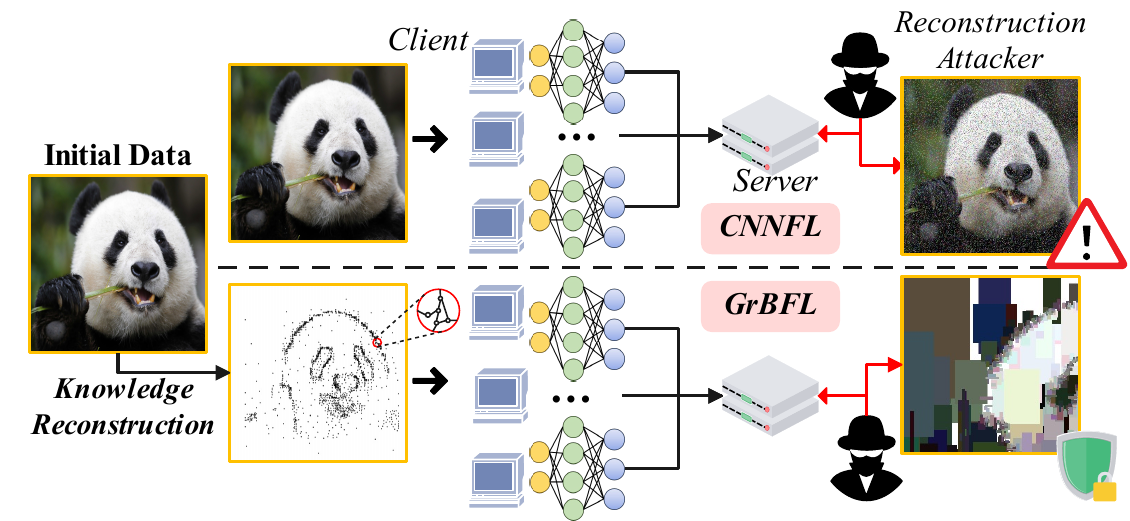}
    \caption{Comparison between the CNNFL and GrBFL processes. The upper section illustrates CNNFL, where image data is directly input into a convolutional neural network, making it vulnerable to attacks. The lower section shows GrBFL, which first reconstructs the image data into a graph before inputting it into a graph neural network, thereby effectively enhancing privacy protection. 
    }
    \label{fig: intro}  
\end{figure}

Significant challenges persist in FL concerning privacy, efficiency, and utility. FL models are vulnerable to various attacks \cite{park2023feddefender,liu2024vertical,rodriguez2023survey,al2023untargeted}, such as those targeting the central server, local devices, or initiated by any participant within the FL workflow. Additionally, the distributed nature of training data across numerous clients, coupled with often unreliable and slow network connections, makes communication efficiency crucial \cite{bao2023communication}. In each learning round, clients independently compute model updates based on their local data and send these updates to the central server, which aggregates them to generate a new global model.

While many studies have addressed issues related to privacy, efficiency, and utility, the "No Free Lunch" (NFL) \cite{zhang2022no,zhang2023trading} principle highlights the lack of a solution that simultaneously improves all three aspects. We approached this problem from the perspective of input data, theoretically analyzing and experimentally validating reconstruction attacks. Our findings indicate that as the information content of input data decreases, the amount of information an attacker can reconstruct also diminishes. An analysis of the expected generalization error suggests that, provided a suitable model is selected and redundant information is removed, the model's classification performance can be maintained. Furthermore, the reduction in information content significantly enhances FL efficiency. Therefore, reconstructing input data and eliminating redundant information presents an effective solution for improving FL.

Building on these findings, we realized that the concept of granular balls can be effectively applied to enhance privacy in FL by reconstructing input data rather than directly using the finest-grained data. Motivated by this insight, we proposed the \textbf{Gr}anular-\textbf{B}all \textbf{F}ederated \textbf{L}earning (GrBFL)  framework and developed a two-dimensional binary search segmentation algorithm based on variance constraints. This framework is designed to detect gradient variations within images, extract representative information, and reconstruct image data into graph structures during the input stage. GrBFL not only ensures that crucial information is preserved with minimal impact on classification accuracy but also enhances privacy by preventing attackers from fully reconstructing the original data due to the loss of some redundant information. Additionally, GrBFL improves the efficiency of FL.

As shown in Figure \ref{fig: intro}, the traditional approach is to transform an image into a matrix and input each pixel to CNN, whereas, in granular-ball input, an image can be represented as a graph after granular-ball computing, where each node represents a structural block in the image and each edge represents the association between two nodes. Local models are trained in each client by GNN and then uploaded to the server for aggregation into global models. The node features contain information about the location and size of the granular rectangles, so the attacker needs to know not only the sample gradient and model parameters but also the meaning of the node features. We simulate an attacker performing a reconstruction attack, showing that the reconstructed image of GrBFL is less similar to the original data than CNNFL. Our model clearly provides better privacy protection and masks partial information of the input image.

In summary, the contributions of this paper are as follows:

\begin{itemize}
    \item[$\bullet$] We rigorously analyze the privacy, efficiency, and utility of FL from the perspective of input and validate our conclusions through experiments. To our knowledge, this is the first method to address the issues of FL from the input perspective.   
    \item[$\bullet$] We propose a novel GrBFL framework based on granular ball generation, which comprises two key components: (1) \emph{Knowledge reconstruction based on granular ball computing}, selecting representative information through gradient analysis and reconstructing it into graph structures. (2) \emph{Joint aggregation based on graph inputs}, introducing a proximal term to address instability issues in graph-based federated learning. Additionally, we designed a two-dimensional binary search segmentation algorithm based on variance constraints to facilitate knowledge reconstruction within this framework.
    \item[$\bullet$] We explored the design of efficiency and comprehensive metrics in federated learning, introducing \emph{Communication Efficiency (CE)} and \emph{Privacy-Efficiency-Utility-Measure (PEUM)} to evaluate the effectiveness of our approach. Extensive experiments demonstrate that our proposed framework significantly enhances efficiency and effectively protects privacy, without compromising practicality.
\end{itemize}

\section{Related Work}

\subsection{Federated learning}
Federated learning (FL) is an emerging collaborative training paradigm that enables participants to train models while preserving data privacy jointly \cite{yang2020federated, gong2020survey}. As FL continues to evolve, the importance of privacy protection has become increasingly prominent. Effectively safeguarding participants' data privacy has become a central research focus, leading to continuous exploration and refinement of privacy-preserving techniques in existing methods \cite{zhang2023trading}.

Currently, privacy protection methods in FL primarily include differential privacy \cite{le2013differentially, chen2024differentially, jiang2024data}, secure multi-party computation \cite{bonawitz2017practical, bell2020secure}, and homomorphic encryption \cite{aono2017privacy, xu2020privacy}. These methods introduce innovations from the perspectives of models and features, enhancing privacy protection capabilities. However, few methods consider privacy protection from the perspective of \textbf{input}. Our research is the first to address this gap and propose a methodological approach in this area.

Moreover, performance and efficiency are equally crucial in FL \cite{zhang2022no, zhang2023trading}. Although, due to the "No-Free-Lunch" principle, achieving significant improvements in all three aspects—privacy, performance, and efficiency—simultaneously is challenging, striving for a balance among them is essential. While previous works have made progress in specific areas, such as FedProx \cite{li2020federated} and FedOPT \cite{ahmed2024fedopt} mitigating performance degradation due to heterogeneity and Scaffold \cite{karimireddy2020scaffold} and FedNova \cite{wang2020tackling} improving efficiency and speed, these advancements remain incomplete. Our proposed framework aims to enhance efficiency while minimizing the loss in accuracy, thereby providing models with stronger generalization capabilities.

\subsection{Granular Ball Computing}

Granular Ball Computing (GBC) is a novel approach within the domain of multi-granularity computing, widely applied in machine learning and data mining. Unlike traditional methods that rely on the finest granularity of input data, such as pixel-level inputs for neural networks, GBC aligns more closely with the way the human brain processes information. Chen's experiments demonstrated that in image recognition, the human brain prioritizes large-scale contour information over fine-grained details \cite{chen1982topological}. This suggests that traditional machine learning models, which use the finest granularity inputs, are not only more susceptible to noise but also less interpretable, and misaligned with human cognitive processes.

Wang first introduced the concept of large-scale cognitive rules into granular computing, leading to the development of multi-granularity cognitive computing \cite{wang2017dgcc}. Building on this, Xia et al. proposed Granular Ball Computing, where different sizes of hyperspheres represent varying levels of granularity: large granular balls for coarse granularity and small balls for fine granularity \cite{xia2020fast, xia2022efficient}.
Recent advancements have further refined this approach. \cite{shuyin2023graph} introduced an enhanced granular ball method for handling non-Euclidean spaces and capturing fine-grained boundary details. By using graph nodes to represent rectangular image regions instead of individual pixels, this method significantly reduces the amount of input data while preserving essential information. This not only enhances efficiency by minimizing redundant data processing but also strengthens privacy protection. 

This deeply prompts us to consider the potential of GBC in enhancing privacy protection within FL. By focusing on coarse-grained and abstract representations rather than fine-grained data points, GBC naturally reduces the exposure of sensitive information. This characteristic presents a critical advantage in privacy-sensitive contexts such as FL.

\section{Understanding Input in Federated Learning}

Existing FL methods mostly focus on improving models and features, with few methods addressing the input level of the model. In this section, we will analyze the relationship between input and privacy, efficiency, and utility in FL from both theoretical and experimental perspectives, yielding practical results.
\subsection{Problem Definition}

FL is a machine-learning 
scenario with $K$ clients (denoted as $\mathcal{C}=\{C_1, C_2, \ldots, C_K\}$), and a central server (denoted as $S$). Each client has its private dataset $\mathcal{D}_i$. The goal of the entire system is to minimize the sum of the loss functions on all clients:
\begin{equation}
    \label{definition}
    \min_w \sum_{k=1}^K \frac{N_k}{N} \cdot \mathcal{F}(w;\mathcal{D}_k)
\end{equation}
where, $ N= \sum_ {k=1}^ K N_k$ is the total number of samples on all clients.

\subsection{Input and Privacy}

A common attack in FL is the data reconstruction attack \cite{zhu2019deep}, which reconstructs the original training data using the model's weights and gradients with the L-BFGS algorithm, posing a significant data security risk. 

The loss function $\mathcal{L}(x, x'; \mathcal{F}) = ||\nabla \mathcal{F}(x) - \nabla \mathcal{F}(x')||^2$ is the optimization objective for the reconstruction attack. This loss function is typically used to measure the difference between the original data $x$ and the reconstructed sample $x'$ and can be a $L_p$ norm or other forms of divergence measures. We use $\mathcal{F}$ to represent the network during the attack process. Consider the impact of a small change $dx'$ in $x'$ on the loss function $\mathcal{L}(x',x'+dx'; \mathcal{F})$:
\begin{equation}
    \label{smalldiff}
    \Delta \mathcal{L} \approx J(x')dx'
\end{equation}
where $\Delta \mathcal{L}$ represents the change in the loss function and $J(x')$ represents the Jacobian matrix of the loss function $\mathcal{L}$ with respect to $x'$.

\begin{wrapfigure}{r}{0.5\textwidth}
    \centering
    \includegraphics[width=1.0\linewidth]{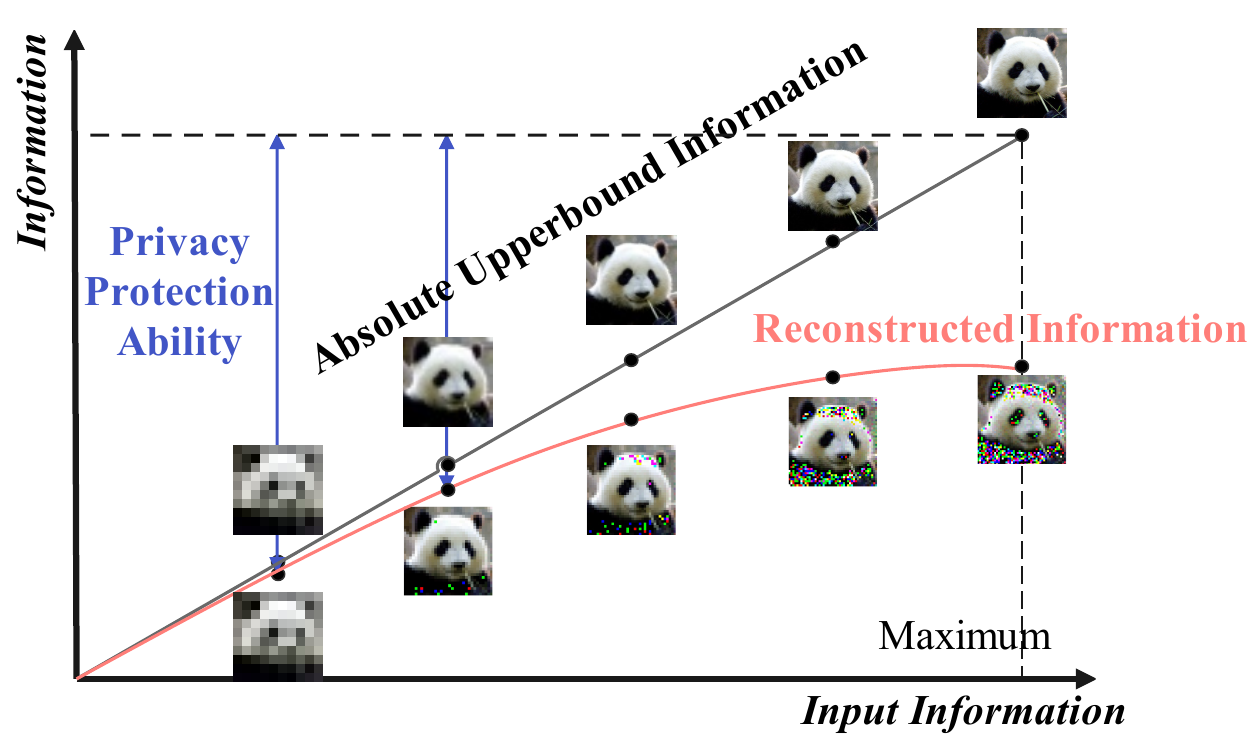}
    \caption{The Relation Between Input and Privacy}
    \label{fig:input_and_privacy}
    \vspace{0.5cm}
\end{wrapfigure}

During the iteration process, $x'$ is updated as follows:
\begin{equation}
    x'_{\text{new}} = x'_{\text{old}} - \alpha \cdot H^{-1} \cdot g
\end{equation}
where $H$ is an approximation of the Hessian matrix, $g$ is the gradient of the current inferred data $x'$, and $H^{-1}$ is estimated using the changes in parameters and gradients over the past few iterations. This approximation better reflects the local curvature information of the current parameters. 

We want the loss function to decrease. Thus, we need:
\begin{equation}
    \label{privacy}
    \Delta \mathcal{L} = - \alpha \cdot g^T \cdot dx' < 0
\end{equation}

The L-BFGS algorithm updates parameters by estimating the inverse of the Hessian matrix. At each iteration, it approximates the inverse matrix of the Hessian at the current parameter value $x'$. Therefore, the direction of $dx'$ is influenced by the inverse of the Hessian matrix. As the input data decreases, the magnitude of $dx'$ becomes smaller. This means that the influence of $dx'$ in the algorithm weakens, and parameter updates are more influenced by the inverse of the Hessian matrix. The L-BFGS algorithm can more accurately estimate the direction of parameter updates, leading to faster convergence to the minimum of the loss function.

Figure \ref{fig:input_and_privacy} illustrates the trend of the information of data reconstructed by the model as the input information changes, which also validates the theory we proposed above. When input data decreases, the reconstruction attacker can better restore the data, but the absolute information content of the obtained data will not exceed that of the original image. Therefore, considering FL's privacy protection from the input perspective is feasible.

\subsection{Input and Utility \& Efficiency}

We can decompose the expected generalization error of FL, analyzing the relationship between input and utility. For a given model $\mathcal{F}$ and data $\mathcal{D}$, the expected generalization error of FL can be expressed as:

\vspace{-0.5cm}
\begin{equation}
\label{expected_error}
    \begin{aligned}
        E(\mathcal{F};\mathcal{D}) &= \sum_{k = 1}^K \frac{N_k}{N} E_{\mathcal{D}_k}[(\mathcal{F}_k(x;\mathcal{D}_k) - y_{\mathcal{D}_k})^2]\\
        &= \sum_{k = 1}^K \frac{N_k}{N} (E_{\mathcal{D}_k}[(\mathcal{F}_k(x;\mathcal{D}_k) - E_{\mathcal{D}}[\mathcal{F}_k(x;\mathcal{D}_k)])^2] + E_{\mathcal{D}_k}[ E_{\mathcal{D}_k}[\mathcal{F}_k(x;\mathcal{D}_k)] - y_{\mathcal{D}_k})^2])\\
        &= \sum_{k = 1}^K \frac{N_k}{N} (\underbrace{E_{\mathcal{D}_k}[(\mathcal{F}_k(x;\mathcal{D}_k) - E_{\mathcal{D}}[\mathcal{F}_k(x;\mathcal{D}_k)])^2]}_{\textbf{Model Bias}}+ \underbrace{E_{\mathcal{D}_k}[( E_{\mathcal{D}_k}[\mathcal{F}_k(x;\mathcal{D}_k)] - y)^2]}_{\textbf{Variance}} + \underbrace{E_{\mathcal{D}_k}[(y_{\mathcal{D}_k} - y)^2]}_{\textbf{Data Noise}}).
    \end{aligned}
\end{equation}

In Equation \ref{expected_error}, we decompose the generalization error into model bias, variance, and data noise. Model bias depends on the model's fitting capability and the proportion of effective information. As the amount of ineffective data decreases, both variance and data noise decrease, while model bias remains unchanged. Conversely, when the amount of effective data decreases, variance and data noise also decrease, but model bias increases. This suggests that it is feasible to retain only the effective information from the input data and reconstruct it, as long as the model's fitting capability is preserved, thereby maintaining the utility of FL.

From an efficiency perspective, for a fixed model and algorithm, when the information content of the input data decreases, the total computation required by the model reduces, thereby enhancing the efficiency of FL.

\section{Methodology}

\subsection{Overview}
\begin{figure*}[!htbp]
  \centering
  \includegraphics[width=0.8\linewidth]{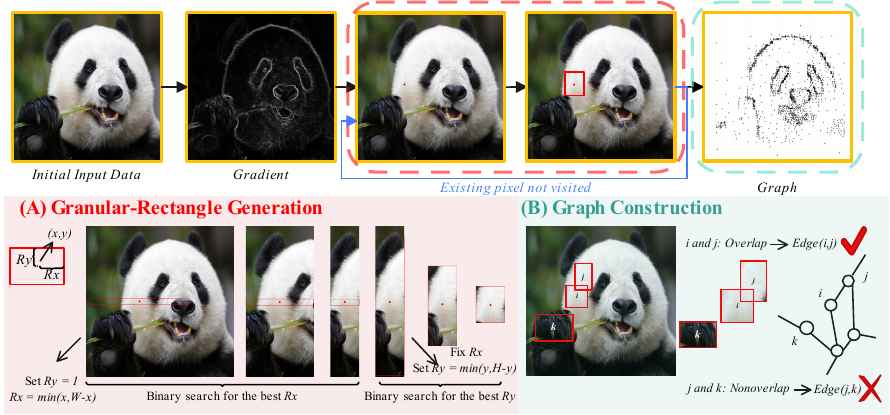}
    \caption{The flowchart presents the granular-ball-based knowledge reconstruction algorithm. The upper section provides an overview of the entire process, while the lower section delves into the technical details of the \emph{Granular-Rectangle Generation} (A) and \emph{Graph Construction} (B) steps.}
  \label{GrB}
  
\end{figure*}

Our proposed framework comprises two main components: \emph{granular-ball-based knowledge reconstruction} and \emph{graph input-based aggregation with variance constraints}. For the knowledge reconstruction, we developed a two-dimensional binary search algorithm based on variance constraints. Figure \ref{GrB} illustrates the knowledge reconstruction process of this algorithm. For the input data, we first compute the gradient map using the \emph{Sobel} operator. Next, we iteratively select the point with the smallest gradient that has not yet been chosen to generate a granular-rectangle, continuing this process until all points have been processed. During the generation of each granular-rectangle, we first fix its height and determine the appropriate length using a bisection method based on purity constraints. Then, with the length fixed, we similarly determine the appropriate height. When no further granular-rectangle can be generated, we reconstruct all the granular information into a graph: each granularity becomes a node, and an edge is formed between any two granular-rectangle that share overlapping pixels. In this way, the image is reconstructed into a graph. We then apply an appropriate graph neural network for federated learning, and perform model aggregation based on variance constraints, before finally uploading the aggregated results to the server.

\subsection{Granular-Ball Based Knowledge Reconstruction}

Equation \ref{privacy} and Equation \ref{expected_error} inspired us to consider how to balance privacy, efficiency, and utility in FL from the input perspective. From the perspectives of utility and efficiency, the performance of a model often depends on the amount of information in the data itself rather than the form in which the data is stored. This insight led us to consider privacy protection from the standpoint of data storage forms. Simply adding noise or blurring the original images can interfere with machine classification accuracy, but reconstructed images by such methods are often still recognizable to humans.

As shown in Figure \ref{fig:input_and_privacy}, with very little information, the model cannot effectively learn the features of a panda, but for humans, it is still recognizable. Therefore, we need to consider filtering and reconstructing the input data to ensure that the reconstructed images are difficult for both humans and machines to recognize. The granular computing methods discussed in \cite{xia2019granular} and \cite{shuyin2023graph} focus on extracting and integrating input information, which is then fed into the model at a coarser granularity. This approach aligns with our concept of reconstructing input information to achieve privacy protection in federated learning. Building on this, we have designed an algorithm for the GrBFL framework.

To achieve image segmentation from fine to coarse granularity, a straightforward approach is to group adjacent similar pixels into the same granular-ball. This allows us to retain the original image structure while removing redundant information. It is important to note that the term "granular-ball" here does not literally refer to a spherical shape, as the image itself is two-dimensional. Therefore, using a rectangle is more appropriate for modeling regions within the image. We first compute the gradient map of the image using operators such as Sobel. For each pixel, the greater the gradient, the fewer similar pixels are typically found in its vicinity. Therefore, we sort the pixels in ascending order of their gradient values and construct granular-rectangle for each pixel accordingly. Purity is a classical metric for assessing the quality of granular-ball generation, which we extend to the context of image granular-rectangles. In image processing, the purity of a granular-rectangle is defined as the ratio of normal pixels to the total number of pixels within the granular-rectangle. Here, normal pixels refer to those whose gray-scale values differ from the granule's center pixel by no more than a specified threshold. The calculation of purity is shown in Equation \ref{purity}.
\begin{equation}
\label{purity}Purity_k=1-\frac{\sum_{i\in R_k}\mathbb{I}(f(p_i)-f(c_k)>thr)}{(2r_x+1)(2r_y+1)},
\end{equation}
where $k$ is the ID of granular-rectangle, $p_i$ is the centre,  $c_k$ is the other pixel points within the granular-rectangle, $thr$ is the defined threshold, $r_x$ and $r_y$ are the length and width of the granular-rectangle, $R_k$ is the granular-rectangle, $f(\cdot)$ is gray-scale value.

Identifying the largest granular rectangle that satisfies the purity constraint is a critical challenge that requires careful consideration. Evaluating each pixel individually can lead to prohibitively long processing times, especially when dealing with large-scale image data. However, it is noteworthy that the purity function demonstrates global monotonicity: as the granular rectangle expands in any direction, its purity consistently decreases. By leveraging this property, a binary search strategy can be employed, starting with a larger size and progressively narrowing down, to efficiently determine the optimal dimensions of the granular rectangle that satisfy the purity constraint. This approach significantly reduces the algorithm's complexity to a logarithmic scale.

After obtaining all the granular rectangles, we construct a graph structure by abstracting the center of each granular rectangle as a node in the graph. Additionally, we consider the pairwise overlapping relationships between granular rectangles. If two granular rectangles overlap, an edge is added between the corresponding nodes; otherwise, no edge is present between the nodes. This graph construction method preserves the original structural information while also encapsulating the relationships between granular rectangles. Formally, the reconstructed graph is defined as:

\begin{definition} (Granular-Rectangle Graph)
    The Granular-Rectangle Graph can be defined as an undirected graph \( G = (V, E) \), where \( V = \{v_1, \ldots, v_n\} \) is the set of granular-rectangle. Each granular-rectangle \( v_k \) is defined as \( v_k = [c_x^k, c_y^k, v_m^k, v_{var}^k, r_x^k, r_y^k, v_{max}^k, v_{min}^k] \), where \( c_x^k \) and \( c_y^k \) represent the coordinates of the granular-rectangle's center, \( r_x^k \) and \( r_y^k \) represent the dimensions of the granular-rectangle, and \( v_m^k \), \( v_{var}^k \), \( v_{max}^k \), and \( v_{min}^k \) denote the mean, variance, maximum, and minimum pixel values within the granular-rectangle, respectively. The edge set \( E \) is defined such that an edge \( (v_i, v_j) \in E \) exists if and only if \( |c_x^j - c_x^i| - 1 < r_x^j + r_x^i \) and \( |c_y^j - c_y^i| - 1 < r_y^j + r_y^i \).
\end{definition}

In the FL scenario, each client processes the input image data as described above, resulting in a series of graph data. The GrBFL framework then inputs these graph data into the graph neural network for training. After a certain number of training rounds, the models from each client will be aggregated. The specific process of model aggregation will be detailed in the following section.

\subsection{Graph Input-Based Aggregations}

Through extensive experiments, we found that, compared to traditional graph datasets like Cora, the datasets generated from granular-ball computing exacerbate the differences between client models in FL. This increased disparity poses challenges for methods that rely on directly weighted averaging of model parameters and gradients, such as slower convergence rates. As shown in Table \ref{table:aggr}, various FL aggregation strategies were evaluated on our dataset, with the results indicating that the simple FedProx \cite{li2020federated} aggregation strategy performed better. The proximal term introduced in FedProx effectively mitigated the instability of graph neural networks when processing granular-ball inputs. The FedProx method is an evolution of the FedAvg \cite{mcmahan2017communication} algorithm, which reduces discrepancies in model parameter adjustments by incorporating a proximal term into the loss function of the client models. This not only accelerates the convergence of the global model but also alleviates the instability encountered by graph neural networks in the context of granular-ball inputs.

\begin{table}[!htbp]\small
    \caption{Aggregation Method Comparison}
    \label{table:aggr}
    \centering
    \begin{tabular}{ccccc}
        \toprule
        \textbf{Method} & \textbf{FedAVG} & \textbf{SCAFFOLD} & \textbf{FedNova} & \textbf{Fedprox} \\
        \midrule
        Acc & 92.33\% & 90.58\% & 90.08\% & \textbf{96.25\%} \\
        \bottomrule
    \end{tabular}
\end{table}

The FedProx algorithm locally introduces a proximal term to limit parameter updates, i.e. $F_{k}(\cdot)$ becomes $h_k(\cdot)$, thus minimizing the following loss function:
\begin{equation}
\min_wh_k(w;~w^t)=F_k(w)+\frac\mu2\|w-w^t\|^2. 
\label{con:loss}
\end{equation}

Following aggregation under the FedProx mechanism, the server transmits the graph neural network model back to the client, thereby concluding a round of FL. 

\subsection{Complexity analysis.}

The time complexity of our method for processing gradients and sorting is \( O(nm \log(nm)) \), where \( n \) and \( m \) represent the length and width of the image, respectively. When solving for the granular-rectangle size using a binary search algorithm, the time complexity is \( O(k \log n \log m) \), where \( k \) is the number of regions and $k \ll nm$. Constructing the granular-rectangle graph has a time complexity of \( O(k^2) \). This method is highly efficient for large-scale image processing. Testing shows that for common 224x224 image data, our method completes data reconstruction in just $1ms$, thereby ensuring model performance.

\section{Experiment}

\subsection{Experiment Setup}

\textbf{Datasets.} We conducts a detailed analysis on three classic image datasets: MNIST, CIFAR-10, and CIFAR-100. The varying difficulty levels of these datasets allow for a comprehensive evaluation of the model's overall performance.

\noindent 
\textbf{Baselines.}  To highlight the strong performance of GrBFL in terms of privacy protection, we compared it with several methods under the CNNFL framework. Specifically, we compared GrBFL with the differential privacy \cite{le2013differentially} and the LotteryFL \cite{li2020lotteryfl} within the CNNFL framework. The former achieves privacy protection from a \emph{feature} perspective, while the latter does so from a \emph{model} perspective.

\noindent 
\textbf{Implementation.} The code for this study was implemented using PyTorch, ensuring that all FL models were evaluated in a consistent computational environment. Within the GrBFL framework, we explored the applicability of this approach across different graph neural network models, including GCN, GAT, GIN, and GraphSAGE. The entire training process was conducted on an NVIDIA RTX-3090 GPU, with specific training parameters and details provided in the supplementary materials.
\subsection{Evaluation Metrics}

To conduct a thorough evaluation of the model within the context of FL, which includes the aspects of privacy, efficiency, utility, and their comprehensive considerations, we carefully define the following metrics.

\noindent \textbf{Privacy.} A key challenge in FL privacy protection is that attackers may infer information about the training data through model parameters and gradients. In this experiment, we assume a scenario where an attacker has successfully obtained model parameters and gradient data related to specific samples from a client. Additionally, we assume that the attacker has comprehensive knowledge of the node feature semantics in the granular computing domain. With this information, the attacker might ultimately reconstruct an image. To evaluate the level of privacy protection achieved under these specific conditions, we introduce a privacy score, denoted as \( S_p \), which serves as a quantitative metric. The formula for the privacy score \( S_p \) is defined in Equation \ref{con:pr}.

\begin{equation} S_p(img_{true},img_{pred}) = 1 - \frac{1}{1 + MSE(img_{true},img_{pred})}, \label{con:pr} \end{equation} where $img_{true}$ refers to the actual image, $img_{pred}$ refers to the image reconstructed by the attacker, and $MSE$ is mean square error.

\noindent \textbf{Efficiency.} We use model communication efficiency (CE) to evaluate the efficiency of different models. In the evaluation of communication efficiency within the FL framework, we meticulously recorded the duration and traffic metrics of communications. Using this data, we calculated the average communication time, which serves as a quantitative indicator of the model's communication efficiency throughout the entire FL cycle. The formal definition of communication efficiency (CE) can be found in Equation \ref{con:ce}.

\begin{equation} CE = 2\sigma( - \frac{\phi \cdot time}{\emph{traffic}}), \label{con:ce} \end{equation} where, \(\sigma(\cdot)\) denotes the sigmoid function, which is a crucial component in calculating communication efficiency. The variables \emph{time} and \emph{traffic} represent the duration of communication and the volume of data communicated, respectively. Specifically, \emph{traffic} is quantified as the product of the number of model parameters and the communication frequency. Additionally, \(\phi\) is a scaling factor used to uniformly map the metric values to a range between 0 and 1.

\noindent \textbf{Utility.} We selected accuracy, a widely recognized multi-class classification metric, as the primary tool for evaluating model utility. Accuracy is used to assess the performance of the aggregated FL model across all test samples, reflecting the aggregation level and representational capability of the individual client models.

\noindent \textbf{Comprehensive measurement.} The evaluation of federated learning (FL) is not limited to a single aspect such as privacy, efficiency, practicality, or isolation. Considering the diversity of these dimensions, we have developed a comprehensive evaluation metric called PEUM (Privacy-Efficiency-Utility Metric) to integrate these three key dimensions and assess the overall effectiveness of the methods considered. The composite score of the model is represented as the harmonic mean of accuracy, communication efficiency, and privacy score. Formally, this is expressed by Equation \ref{con:com}.

\begin{equation} \emph{PEUM} = \frac{1}{\frac{1}{Acc} + \frac{1}{CE} + \frac{1}{S_p}}. \label{con:com} \end{equation}

\subsection{Main Results}

\textbf{The GrBFL framework demonstrates exceptional efficacy in privacy protection.} 
\begin{wrapfigure}{r}{0.6\textwidth}
    \centering
     \includegraphics[width=\linewidth]{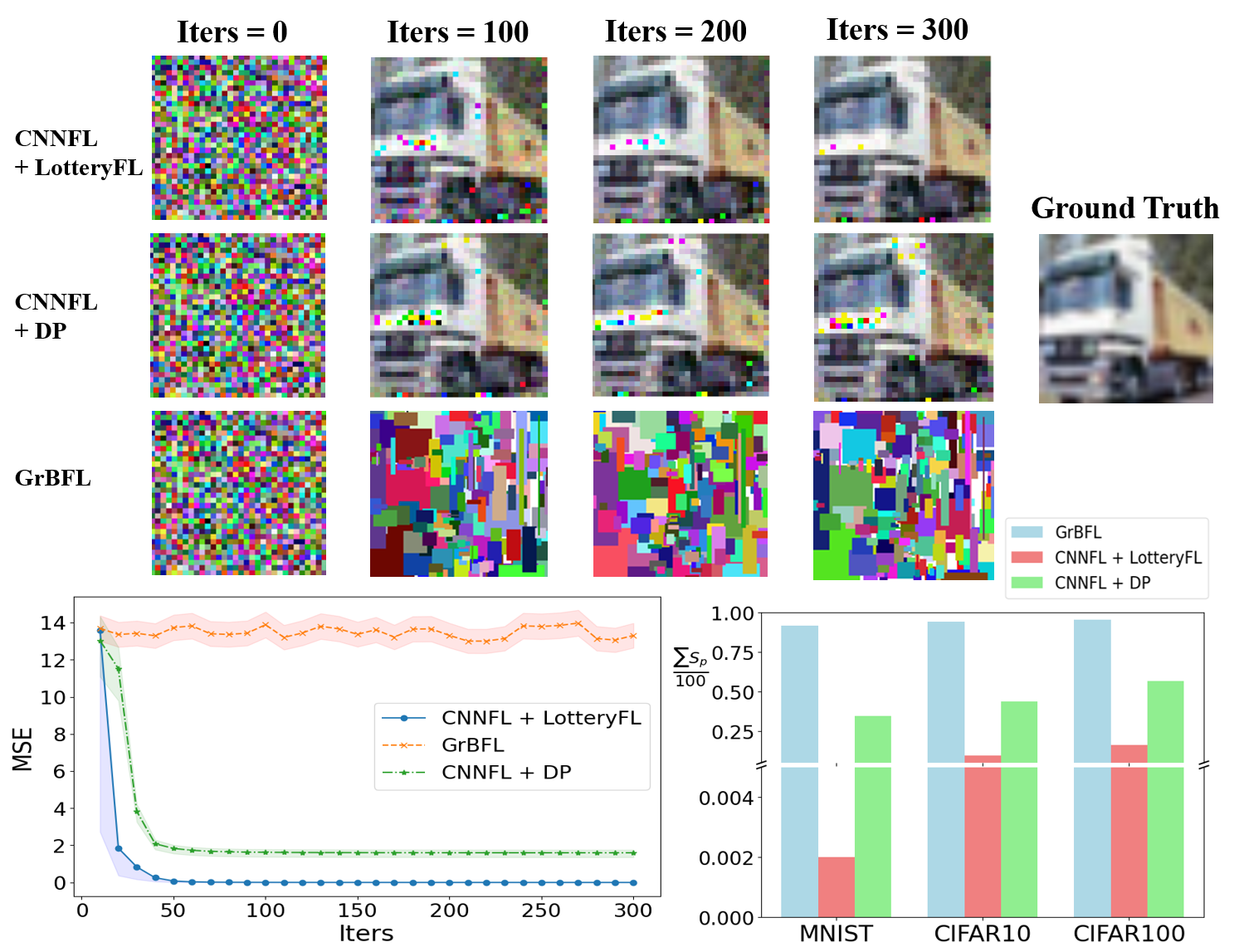}
    \caption{Privacy Protection Experiment.}
    \label{fig:sp}
\end{wrapfigure}
To rigorously evaluate the privacy protection capabilities of the GrBFL and CNNFL frameworks, this study conducted reconstruction attack experiments under the GCN and ResNet models. The experiments were performed across three datasets. For each dataset, a set of 100 random training samples was systematically extracted and subjected to 300 iterations to compute $S_p$.
The attack model for CNNFL can be referenced in \cite{zhu2019deep}. The attack model for GrBFL is similar to that of CNNFL. In this case, the attacker can access the weights of the graph neural network and the gradients of the data, using continual optimization to reconstruct the original graph data. We assume that the attacker understands the semantics of node features. However, since node features only contain coarse pixel information, the attacker can only determine the position and size of each granule rectangle but cannot precisely obtain the specific pixel values.

Figure \ref{fig:sp} illustrates the results of our privacy protection experiments. The upper section compares the differential privacy performance of GrBFL, LotteryFL, and Differential Privacy (DP) across different numbers of iterations. It can be observed that, although LotteryFL and DP provide some degree of privacy protection, reconstruction of the data remains possible as the number of iterations increases. In contrast, GrBFL demonstrates significant effectiveness in privacy protection. Even with a surge in the number of iterations, GrBFL's inherent information loss mechanism effectively hinders the attacker's ability to accurately reconstruct the training data.

Figure \ref{fig:sp} quantitatively displays the results of the privacy protection experiments. The line chart in the lower left quadrant shows the trajectory of mean squared error (MSE) loss between reconstructed images and the original images for different scenarios, indicating that GrBFL maintains a significant divergence in data recovery, thereby validating its effectiveness in privacy protection. The bar chart in the lower right section presents the average \( S_p \) results for 100 random samples from three datasets, demonstrating that GrBFL exhibits strong privacy protection capabilities across all datasets.

\begin{wrapfigure}{r}{0.6\textwidth}
    \centering
     \includegraphics[width=\linewidth]{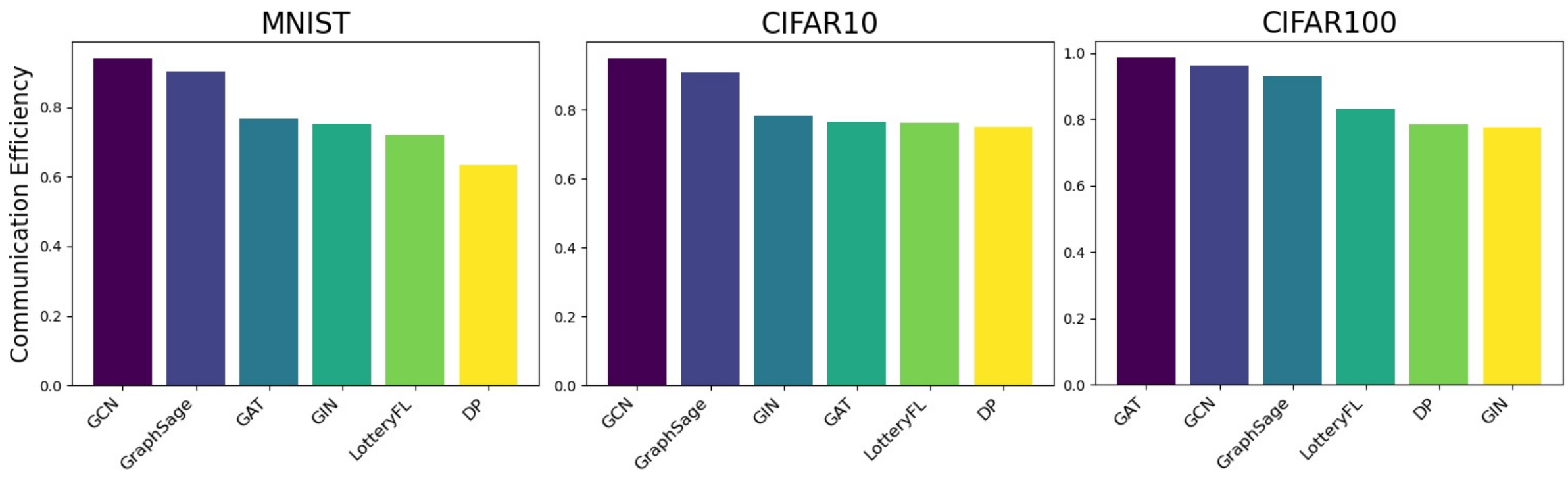}
    \caption{Comparison of communication efficiency}
    \label{fig:cee}
\end{wrapfigure}

\noindent 
\textbf{GrBFL achieves superior communication efficiency.} Figure \ref{fig:cee} provides a visualization of communication efficiency. The experimental results demonstrate that GrBFL performs exceptionally well in terms of communication efficiency. Specifically, in the three experimental datasets, when the parameter is set to \(\phi = 3 \times 10^6\), the model efficiency score for GrBFL is significantly higher compared to FL methods based on the CNNFL framework.

\noindent 
\textbf{GrBFL performs well in the comprehensive evaluation.} We conducted a comprehensive assessment of each model using the metrics defined in Equation \ref{con:com}, with the results shown in Table \ref{tab: coms}. Although privacy protection methods from both feature and model perspectives achieve some degree of privacy protection, their overall scores are not balanced. This validates the rationale of considering knowledge reconstruction from the input perspective, as this approach effectively protects privacy without excessively compromising efficiency and utility, and even achieves better overall performance.

\begin{table}[h]
\caption{Comprehensive evaluation for GrBFL and CNNFL. Among them, the best results are bolded for emphasis.}
\label{tab: coms}
\centering
\setlength{\tabcolsep}{6pt} 
\begin{tabular}{ccccc}
\hline
Framework & Methods & MNIST & CIFAR10 & CIFAR100 \\ \hline
 & \cellcolor[HTML]{EFEFEF}GAT & \cellcolor[HTML]{EFEFEF}0.29 $\pm$ 0.02 & \cellcolor[HTML]{EFEFEF}0.24 $\pm$ 0.05 & \cellcolor[HTML]{EFEFEF}\textbf{0.16 $\pm$ 0.07} \\
 & GCN & \textbf{0.32 $\pm$ 0.02} & \textbf{0.25 $\pm$ 0.03} & 0.14 $\pm$ 0.04 \\
 & \cellcolor[HTML]{EFEFEF}GIN & \cellcolor[HTML]{EFEFEF}0.29 $\pm$ 0.01 & \cellcolor[HTML]{EFEFEF}0.22 $\pm$ 0.06 & \cellcolor[HTML]{EFEFEF}0.12 $\pm$ 0.04 \\
\multirow{-4}{*}{\begin{tabular}[c]{@{}c@{}}GrBFL\\ (Ours)\end{tabular}} & GraphSage & 0.31 $\pm$ 0.00 & 0.22 $\pm$ 0.02 & 0.12 $\pm$ 0.02 \\ \hline
 & \cellcolor[HTML]{EFEFEF}DP & \cellcolor[HTML]{EFEFEF}0.27 $\pm$ 0.04 & \cellcolor[HTML]{EFEFEF}0.21 $\pm$ 0.05 & \cellcolor[HTML]{EFEFEF}0.11 $\pm$ 0.04 \\
\multirow{-2}{*}{CNNFL} & LotteryFL & 0.22 $\pm$ 0.03 & 0.22 $\pm$ 0.01 & 0.12 $\pm$ 0.03 \\ \hline
\end{tabular}
\end{table}

\begin{figure}[!ht]
    \centering
    \includegraphics[width=0.8\linewidth]{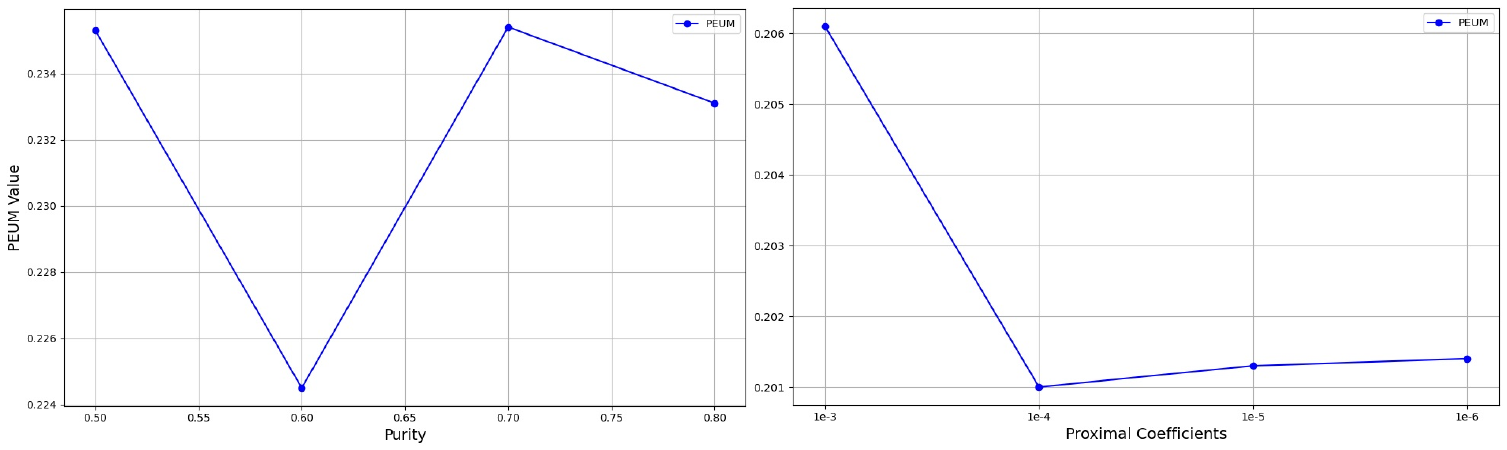}
    \caption{Parameter Sensitivity Study. This experiment was performed on the GAT model using the CIFAR10 dataset.}
    \label{fig: syn}
\end{figure}

\noindent 
\textbf{Parameter Sensitivity Study.} We conducted a parameter sensitivity analysis on the purity of the constructed granular-receptacle and the proximal term coefficient in the client model's loss function. The results in Figure \ref{fig: syn} indicate that our method exhibits strong robustness, with the impact of parameter variations not exceeding 0.1.

Additionally, we conducted experiments on membership inference attacks. The detailed results of additional experiment results, along with the details of experiments mentioned earlier, can be found in the supplementary materials.
\section{Conclusion}
In this paper, we theoretically analyzed the feasibility of achieving a balance between privacy, efficiency, and utility in FL from the input perspective. Based on the concept of granular-ball computation, we proposed a new FL framework called GrBFL. We converted images into graphs through adaptive granular ball computation and then input them into a graph FL model for classification. Additionally, we designed two metrics to evaluate the effectiveness of the proposed FL paradigm. Experimental results validated the correctness of this input perspective approach. In future work, we will continue to investigate the impact of inputs in FL and develop more algorithms for GrBFL.



\end{document}